\documentclass{article}
\usepackage{spconf,amsmath,graphicx}
\usepackage{booktabs}
\usepackage {xcolor} 
\usepackage{url}

\title{High Resolution Image Quality database}
\twoauthors
  {Huang Huang, Qiang Wan\sthanks{Equal contribution with H. Huang.}}
    {Shenzhen University\\
    School of Computer Science and\\ 
    Software Engineering\\
    Shenzhen, P.R. China}
  {Jari Korhonen\sthanks{This work was supported in part by Guangdong ”Pearl River Talent Recruitment Program” under Grant 2019ZT08X603.}}
    {University of Aberdeen\\
     School of Natural and Computing Sciences\\
     Aberdeen, UK\\}

\usepackage{fancyhdr}

\begin{document}

\fancypagestyle{firststyle}
{
   \fancyhf{}
   \fancyfoot[L]{\footnotesize \copyright 2024 IEEE. Personal use of this material is permitted. Permission from IEEE must be obtained for all other uses, in any current or future media, including reprinting/republishing this material for advertising or promotional purposes, creating new collective works, for resale or redistribution to servers or lists, or reuse of any copyrighted component of this work in other works.}
   \renewcommand{\headrulewidth}{0pt} % removes horizontal header line
}

\maketitle
\thispagestyle{firststyle}
\begin{abstract}
With technology for digital photography and high resolution displays rapidly evolving and gaining popularity, there is a growing demand for blind image quality assessment (BIQA) models for high resolution images. Unfortunately, the publicly available large scale image quality databases used for training BIQA models contain mostly low or general resolution images. Since image resizing affects image quality, we assume that the accuracy of BIQA models trained on low resolution images would not be optimal for high resolution images. Therefore, we created a new high resolution image quality database (HRIQ), consisting of 1120 images with resolution of $2880\times2160$ pixels. We conducted a subjective study to collect the subjective quality ratings for HRIQ in a controlled laboratory setting, resulting in accurate MOS at high resolution. To demonstrate the importance of a high resolution image quality database for training BIQA models to predict mean opinion scores (MOS) of high resolution images accurately, we trained and tested several traditional and deep learning based BIQA methods on different resolution versions of our database. The database is publicly available in \url{https://github.com/jarikorhonen/hriq}.
\end{abstract}
\section{Introduction}
\label{sec:intro}

\begin{figure}[ht]
\begin{minipage}[b]{.48\linewidth}
  \centering
  \includegraphics[width=\linewidth]{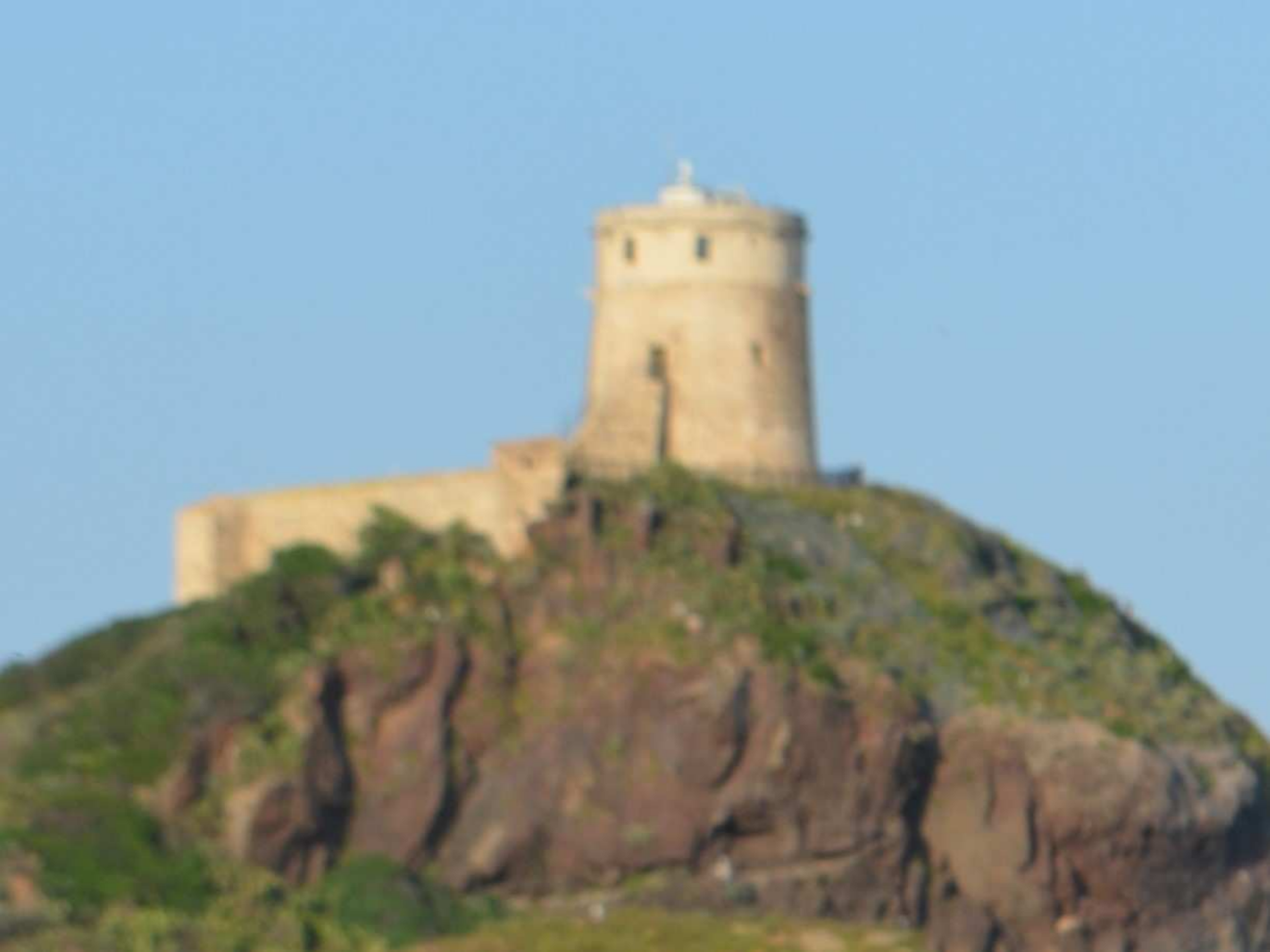}
% \centerline{\epsfig{figure=image3.ps,width=4.0cm}}
%   \vspace{1.5cm}
  \centerline{\small(a) Patch from the original image.}\medskip
\end{minipage}
\hfill
\begin{minipage}[b]{0.48\linewidth}
  \centering
  \includegraphics[width=\linewidth]{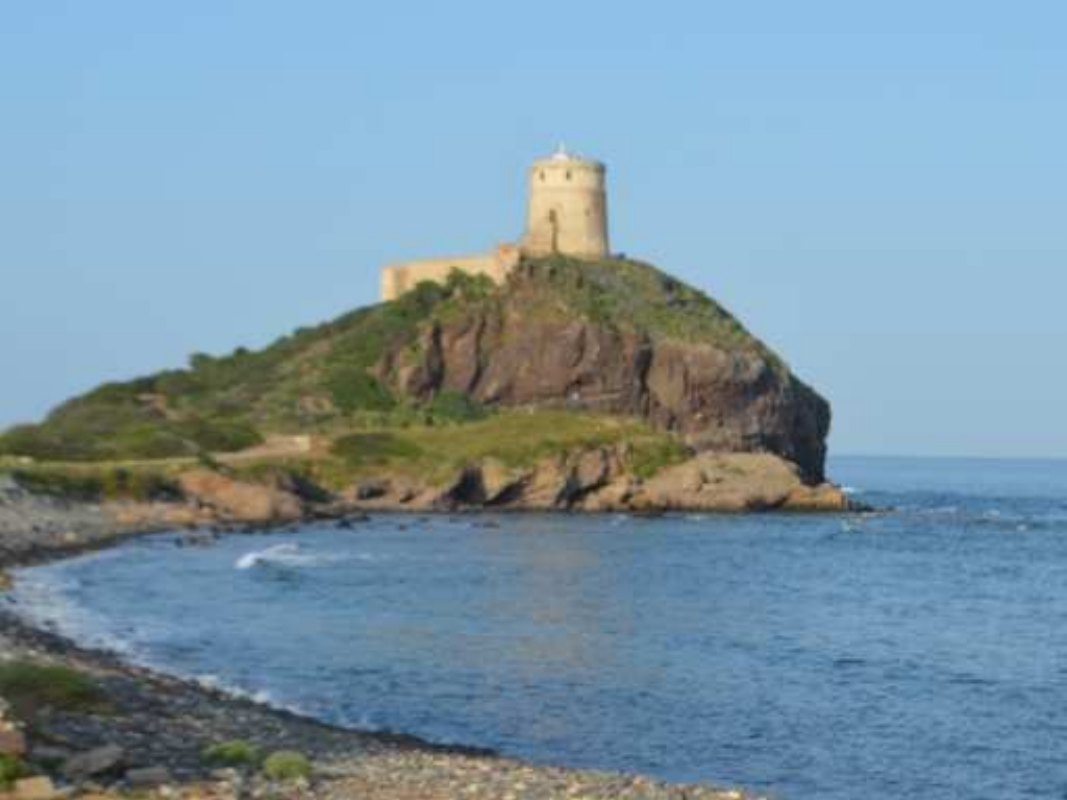}
% \centerline{\epsfig{figure=image4.ps,width=4.0cm}}
%   \vspace{1.5cm}
  \centerline{\small(b) Full image resized.}\medskip
\end{minipage}
\caption{Comparison of the visual effects of blur at high and low resolutions. Figure (a) shows a patch of $512\times384$ pixels cropped from the original image of $2880\times2160$ pixels, and Figures (b) shows the corresponding full image resized to $512\times384$ pixels.}
\label{fig:fig1}
\end{figure}

Image Quality Assessment (IQA) is required to evaluate the perceived impact of distortions in images induced during capture, compression, transmission, and display. In many applications, IQA is essential to optimize the quality of the images presented to the user. In general, IQA methods can be divided in two categories: subjective and objective IQA. In subjective IQA, image quality is assessed by human observers to obtain a subjective quality score, such as MOS, for each image. Subjective IQA is time-consuming and expensive, but since perceived quality is by definition based on human judgement, subjective ratings are necessary to obtain the ground truth MOS~\cite{hosu2020koniq}. In contrast, objective IQA methods aim at predicting the ground truth MOS directly from the image without human involvement. Compared with subjective IQA, objective IQA is more efficient and easier to use. However, subjective IQA is still needed for training, testing and calibrating the objective IQA methods.

Since training of objective IQA methods typically require a large amount of image data annotated with ground truth MOS labels, subjective image quality databases are particularly important. During the past twenty years, a large number of IQA databases have been created. The traditional databases, such as IVC~\cite{le2005subjective}, LIVE~\cite{sheikh2006statistical}, TID2008~\cite{ponomarenko2009tid2008}, CSIQ~\cite{larson2010most}, TID2013~\cite{ponomarenko2015image}, and CID2013~\cite{virtanen2014cid2013} have a limited variety of contents and contain mostly artificial distortions. LIVE-itW~\cite{ghadiyaram2015massive}, published in 2016, is the first large-scale database (over 1,000 images) with authentic in-capture distortions. Recently, even larger databases with over 10,000 images have been published. KoNIQ-10k~\cite{hosu2020koniq} is a natural image quality database with subjective scores collected via crowdsourcing on the internet, and SPAQ~\cite{fang2020perceptual} is an image quality database focusing on smartphone images rated in a lab-based study. Large image quality databases with artificial distortions are also available~\cite{pdap-hdds,lin2019kadid} contains artificially distorted images rated via crowdsourcing.

In this paper, we focus on images with authentic distortions, such as typical consumer photographs taken with smartphones or standard digital cameras. In this type of images, low-level distortions, such as sensor noise or subtle out-of-focus blur, will be easily noticed by human assessors at the original high resolution. However, those distortions will disappear when the image is downsized. This is demonstrated in Fig.~\ref{fig:fig1}, showing a comparison of two images after cropping and resizing, respectively. As the example shows, blurriness is very obvious in the cropped image, whereas the downsized version looks clean and sharp. Furthermore, the resolution and physical size of the display, as well as viewing distance, also impact the perceived quality. Viewing a high resolution image on a small screen with high pixel density, such as high quality smartphone display, has essentially similar effect as image downsizing on a large monitor with standard pixel density. In addition, we assume that the mental process of assessing a high resolution image that does not fit in the human central field of view on a large display differs from the process of assessing a small image occupying only part of the display. For these reasons, we cannot assume that the BIQA models giving accurate results for low resolution images will also give accurate results for high resolution images.

Unfortunately, the resolutions of the images used for acquiring MOS for the publicly available large scale natural image quality databases tend to be relatively low. Most of the images in LIVE-itW database~\cite{ghadiyaram2015massive} have resolution of $500\times500$ pixels only. KoNIQ-10k database uses higher resolution of $1024\times768$~\cite{hosu2020koniq}, which is still well below the standard Full HD display resolution of $1920\times1080$. PDAP-HDDS database~\cite{pdap-hdds} includes $12,000$ images in Full HD resolution, but the distortions have been generated artificially. SPAQ database includes high resolution original images, but low resolution version of the images with the longest side constrained to $512$ pixels were used to collect the subjective ratings~\cite{fang2020perceptual}, and therefore the subjective ratings do not accurately represent the quality of the original full resolution images. Cross-resolution image quality database KonX~\cite{konx} includes high resolution images of $2048\times1536$, but the database comprises only 420 source images. It is also worth noting that the subjective ratings for LIVE-itW, KoNIQ-10k, and KonX databases were collected in the internet, and therefore the results incorporate a mixture of different display devices and viewing conditions. Apparently, there is a demand for a new large scale image quality database with high resolution content with natural distortions, rated in a controlled laboratory environment with a large high resolution display.

In this paper, we aim to fill the gap in high resolution subjective image quality databases and introduce the highest resolution natural image quality database to date. The database consists of 1120 images captured with a variety of devices including standard digital cameras and smartphones. The images were rated by 175 test subjects using a high resolution monitor in a laboratory environment with controlled viewing conditions. To verify the usefulness of the database for training and testing BIQA methods for high resolution images, we experimented several commonly used BIQA methods, representing the state-of-the-art, on different resolution versions of our database ($2880\times2160$, $1024\times768$, $512\times384$). The experimental results support our hypothesis that BIQA models trained and tested with the low resolution version do not achieve optimal performance.

\section{Database Creation}

HRIQ database was created in three stages. First, we manually selected the source material and processed it by cropping and resizing to a fixed resolution. Second, we conducted a subjective quality assessment study to obtain quality ratings for computing MOS for each image. Third, we analyzed the subjective results to remove potential outliers. In this Section, the database creation process is described in detail.

\subsection{Content Collection}
 In this work, our goal was to create a database with typical consumer photos taken with non-professional devices in everyday life for sharing in social media or saving in a private album. To ensure that our database is a relatively accurate representation of real world consumer photos, the images were selected from the private albums of the authors, taken with mainstream capture devices such as Android and Apple phones and standard DSLR cameras. The images have a high diversity of content, including daily life scenes such as buildings, people, vehicles, food, text slogans, etc., as well as natural scenes such as sky, ocean, plants, and animals. The content includes daytime and night scenes, taken under artificial light and different weather conditions outdoors. The dataset is also geographically diverse, as the photos are taken in several different countries and continents. In terms of distortion, the images contain a variety of authentic distortions, including noise, out-of-focus blur, motion blur, overexposure, underexposure, low contrast, incorrect saturation, and combined distortions. Moreover, we have selected a wide range of images with distortions that are easily overlooked at low resolutions, but can significantly affect ratings at high resolutions, such as subtle out-of-focus blur, sensor noise, etc.

The original source images, captured with several different devices, represent a range of resolutions from $4000\times3000$ to $8000\times6000$. The original image format was JPEG, with a mixture of aspect ratios 4:3 and 16:9. Since the resolutions of standard consumer displays are typically much lower, we resized the images to $2880\times2160$. Before resizing, the images with aspect ratio of 16:9 were cropped vertically in the center to obtain 4:3 aspect ratio. We chose the final resolution of $2880\times2160$, because the native resolution of the display used in the study is $3840\times2160$; therefore, the final images would occupy the full height of the display. The remaining area of the screen would be reserved for the user interface. The aspect ratio of 4:3 was chosen as it is the original aspect ratio of most of the source images. The PIL library in Python was used for cropping and resizing the images to retain the highest possible quality in resizing. For testing different BIQA models on different resolutions, we also created $1024\times768$ and $512\times384$ resolution versions of the images; however, only $2880\times2160$ resolution was used for subjective testing.

\subsection{Subjective Quality Assessment Study}
Most of the recently published large-scale image quality databases have been rated by the users in internet-based crowdsourcing studies. However, it is not realistic to expect that the most users would have high resolution displays available. Therefore, our subjective tests were conducted in a lab environment with controlled conditions to ensure that the display device and the viewing conditions allow reliable rating of high resolution images. The test users were recruited at the campus, which means that the test users were all college students. We briefly screened the test users to ensure that they were not color-blind or color-weak, etc. In total, 175 test users participated in the test. The average age of the participants was 22 (ranging from 18 to 26), 70\% of the users were male and 30\% female, 70\% of the participants used glasses, and 11\% of test users had prior IQA-related knowledge and experience. 

\begin{figure}[htbp]
    \centering
    \includegraphics[width=0.48\textwidth]{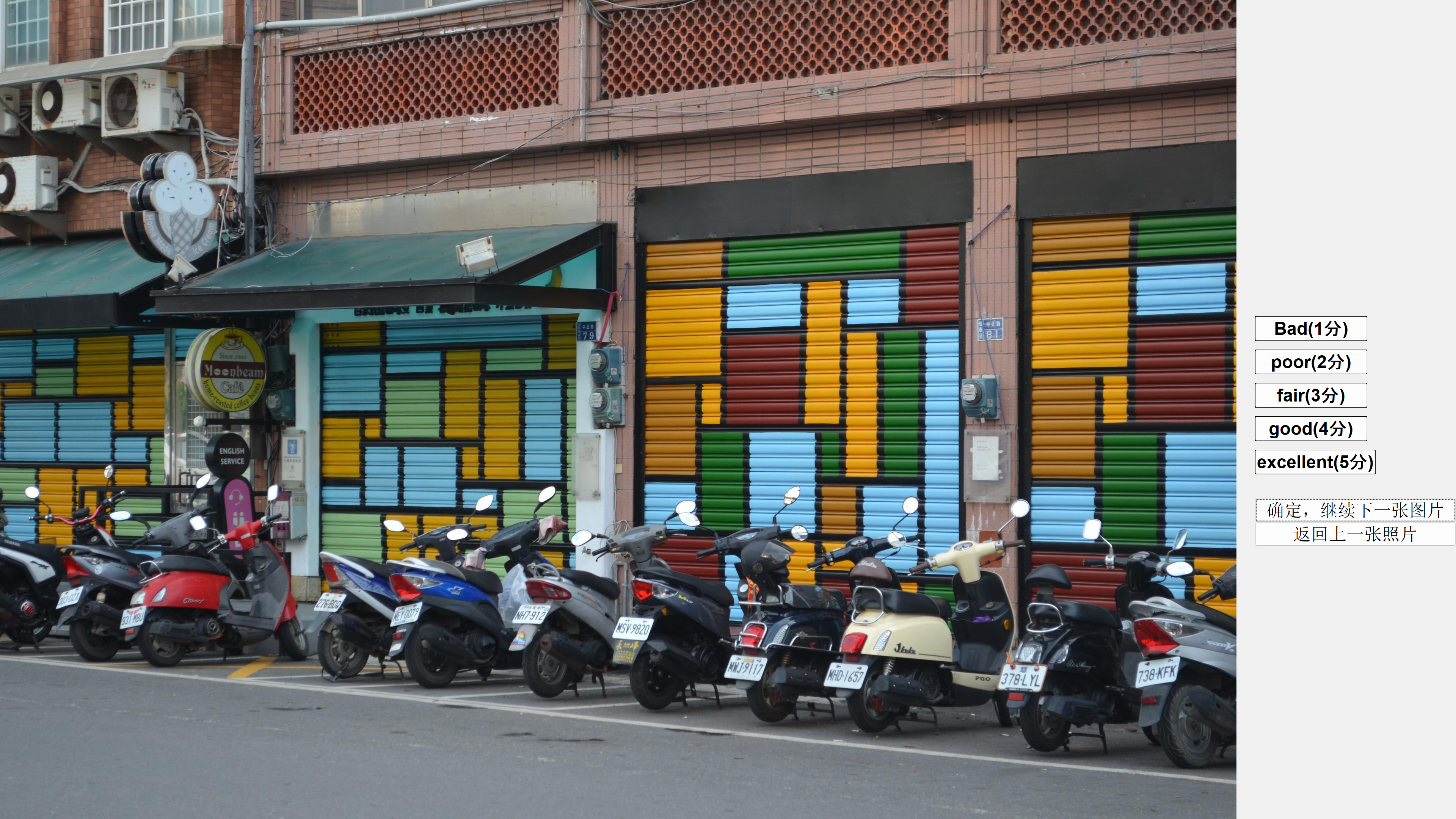}
    
    \caption{Subjective rating interface. The $2880\times2160$ resolution image displayed on a $3840\times2160$ monitor takes most of the screen. The user interface for rating is on the right side of the screen.}
    \label{fig:fig2}
\end{figure}

In the practical test arrangements, we followed the ITU-T guidelines~\cite{itu-t} for visual quality assessment when practically applicable. The test was conducted in a peaceful laboratory room, with four Dell U2720Q 4K monitors of $3840\times2160$ resolution. The lighting environment of the laboratory was conventional. Each of the 175 test users assessed 160 images to ensure that each image would be evaluated by 25 different users. We allowed users to adjust the monitor position and angle for a convenient viewing experience. To avoid testing fatigue to affect the results, we instructed the users to spend approximately 5-10 seconds for assessing each image, to make sure all the assigned images would be evaluated in 15-30 minutes. The interface of the used testing program is shown in Fig.~\ref{fig:fig2}. The image will occupy most of the screen, and the rating buttons are located in a small area on the right side of the screen. The standard five-point absolute category rating (ACR) scale was used in the experiment, i.e. the scores and the respective image quality levels were defined as 1: bad, 2: poor, 3: fair, 4: good, 5: excellent.

%\begin{figure*}[ht]
%\begin{minipage}[b]{.48\linewidth}
%  \centering
%  \includegraphics[width=\linewidth]{tester mean1.png}
% \centerline{\epsfig{figure=image3.ps,width=4.0cm}}
%   \vspace{1.5cm}
%  \centerline{(a) distribution of mean differences}\medskip
%\end{minipage}
%\hfill
%\begin{minipage}[b]{0.48\linewidth}
%  \centering
%  \includegraphics[width=\linewidth]{tester std1.png}
% \centerline{\epsfig{figure=image4.ps,width=4.0cm}}
%   \vspace{1.5cm}
%  \centerline{(b) distribution of standard deviations}\medskip
%\end{minipage}
%
% \begin{minipage}[b]{.48\linewidth}
%   \centering
%   \includegraphics[width=\linewidth]{tester mean2.png}
% % \centerline{\epsfig{figure=image3.ps,width=4.0cm}}
% %   \vspace{1.5cm}
%   \centerline{(c) Gap mean after}\medskip
% \end{minipage}
% \hfill
% \begin{minipage}[b]{0.48\linewidth}
%   \centering
%   \includegraphics[width=\linewidth]{tester std2.png}
% % \centerline{\epsfig{figure=image4.ps,width=4.0cm}}
% %   \vspace{1.5cm}
%   \centerline{(d) Gap std after}\medskip
% \end{minipage}
%
%\caption{The histograms showing the distribution of the mean gap between MOS and the ratings by each individual user (a) and the standard deviations of the gaps (b), respectively. We used this data to find the outliers: histogram (b) shows that one user has substantially larger standard deviation than the others, indicating high inconsistency of the user's ratings in respect with MOS. Therefore, the user was considered as an outlier and removed from the final MOS results.}
%\label{fig:fig3}
%\end{figure*}

\subsection{Subjective Data Analysis}
From the subjective experiment, we obtained 25 ratings for each image, and first calculated the preliminary MOS scores. Although we obtained the results in a controlled laboratory setting, typically giving more reliable ratings than crowdsourcing experiments, outlier users still need to be identified and excluded from the final results. First, for each user, we calculated the differences between the ratings and the respective MOS. Then, for each user, we calculated the mean and standard deviation of the differences between user's ratings and MOS. 

We observed that the distribution of mean differences and their standard deviations roughly resemble Gaussian distribution. Some users seem to give systematically slightly higher or smaller ratings than the others, but this does not necessarily mean that they are unreliable, if the difference is consistent. Therefore, we used standard deviation as our main criterion for detecting outliers. One user showed significantly higher standard deviation than the others, indicating that this specific user had given inconsistently higher and smaller ratings than the other users for the same images. Therefore, the user was excluded from the final MOS results.

 After excluding the outlier, some of the images have only 24 ratings. Since the subjective quality evaluation was performed in a laboratory, we expect the final MOS results to be relatively accurate. The final distribution histogram of the MOS seems relatively even, with a small overrepresentation of images in the quality range from three to four, as well as a small underrepresentation of very high quality images with MOS above 4.5. We assume that the test users were rather conservative for giving full rating of five.
%original place for fig3

%\begin{figure}[htbp]
%    \centering
%    \includegraphics[width=0.48\textwidth]{mos d2.png}
%    
%    \caption{MOS distribution of HRIQ.The distribution is roughly even, at most in the middle %score range, that is, scores of 3 and 4.}
%    \label{fig:fig4}
%\end{figure}

\begin{table*}[ht]
  \centering
  \caption{Performance of the selected BIQA models on the proposed HRIQ database.}
  \vspace*{1mm}
  \setlength{\tabcolsep}{6.5mm}{
  \begin{tabular}{c|cc|cc|cc}
    \toprule
          & \multicolumn{2}{c|}{\textbf{HRIQ2880}} & \multicolumn{2}{c|}{\textbf{HRIQ1024}} & \multicolumn{2}{c}{\textbf{HRIQ512}} \\
\cmidrule{2-7}    \textbf{Method} & \textbf{SROCC} & \textbf{PLCC} & \textbf{SROCC} & \textbf{PLCC} & \textbf{SROCC} & \textbf{PLCC} \\
    \midrule
    \midrule
    DIVINE & 0.381 & 0.422 & \textbf{0.440} & \textbf{0.451} & 0.272 & 0.289 \\
    BRISQUE & 0.063 & 0.177 & \textbf{0.293} & \textbf{0.331} & 0.261 & 0.272 \\
    BIQI  & \textbf{0.559} & \textbf{0.572} & 0.400   & 0.343 & 0.185 & 0.244 \\
    HOSA  & \textbf{0.507} & \textbf{0.520} & 0.475 & 0.487 & 0.414 & 0.434 \\
    \midrule
    DBCNN & - & - & \textbf{0.895} & \textbf{0.899} & 0.856 & 0.863 \\
    Koncept512 & -     & -     & \textbf{0.732}     & \textbf{0.726}     & 0.700 &0.650 \\
    HyperIQA & 0.848  & 0.848 & \textbf{0.873} & \textbf{0.879} & 0.847 & 0.854 \\
    LinearityIQA & -     & -   & \textbf{0.895} & \textbf{0.901}  & 0.846    & 0.859      \\
    MANIQA & 0.824  & 0.824 & 0.884 & 0.891 & \textbf{0.899} & \textbf{0.909} \\
    HR-BIQA & \textcolor[rgb]{ 1,  0,  0}{\textbf{0.920}} &\textcolor[rgb]{ 1,  0,  0} {\textbf{0.925}} & 0.904 & 0.912 & 0.849  & 0.859 \\
    \bottomrule
    \end{tabular}%  
  }
  \label{tab:tab2}%
\end{table*}%

\section{Evaluating BIQA methods on HRIQ}
We evaluated representative BIQA methods on HRIQ with different resolutions, including traditional methods and deep learning-based methods. For the traditional BIQA methods, we selected BIQI~\cite{biqi}, BRISQUE~\cite{brisque}, DIVINE\cite{diivine}, and HOSA~\cite{hosa}, and we tested the HRIQ database using the pre-trained models directly. For the deep methods, we selected the state-of-the-art BIQA models DBCNN~\cite{dbcnn}, HyperIQA~\cite{hyperiqa}, KonCept512~\cite{hosu2020koniq}, LinearityIQA~\cite{line}, and MANIQA~\cite{maniqa}. We also included new high resolution BIQA model HR-BIQA, inspired by our earlier model RNN-BIQA~\cite{rnn}. 

To our knowledge, RNN-BIQA is the only prior BIQA model designed specifically for high-resolution images. It is a patch-based model, using a deep convolutional neural network (CNN) to extract features from patches, and a separately trained recurrent neural network (RNN) to obtain the quality scores from a sequence of feature vectors extracted from the patches. Unfortunately, RNN-BIQA has been tested on relatively low resolution images only, and in our experiments, it did not perform optimally on the HRIQ database. Therefore, we redesigned the model, using a modified ResNet50 CNN fine-tuned for IQA and vanilla vision transformer (ViT) combined for feature extraction, followed by two RNN streams for the full resolution and low resolution versions of the input image for spatial pooling and MOS prediction. Due to the space constraints, detailed description of HR-BIQA is not given here, but the source code and more details of the model are available in~\cite{hriqdatabase}.

%\begin{figure}[htbp]
%    \centering
%    \includegraphics[width=0.48\textwidth]{icmefig5.pdf}
%    \caption{High level illustration of the proposed HR-BIQA model architecture.}
%    \label{fig:fig4}
%\end{figure}

In the comparison study, we randomly divided the HRIQ database into a training set with 80\% of the images and a testing set with 20\% of the images. We trained and tested the models using 24GB RTX3090 GPU, and we repeated the experiments ten times using different seeds to randomly select the training and test sets. Default configuration and hyperparameters provided by the respective authors were used for training the benchmark models. For fair comparison, we used the same partitioning for all the models, as well as different resolution versions of the database. It is worth noting that we were not able to run DBCNN, KonCept512, and LinearityIQA on the full resolution database, because the GPU run out of memory. This highlights the challenges of using BIQA models originally designed for standard images to predict high resolution image quality.

We evaluated the model performance using Spearman rank order correlation coefficients (SROCC) and Pearson linear correlation coefficients (PLCC). The reported results are the averages of the ten random partitions. From the results shown in Table~\ref{tab:tab2} we can see that the traditional BIQA methods do not work well on HRIQ, and the results for the deep methods are substantially better. Concerning all resolutions, HR-BIQA achieves the best overall performance on HRIQ2880 with a clear margin.

The results on different resolutions show that BIQI, HOSA, and HR-BIQA perform the best on the full resolution database, MANIQA shows the best result on the lowest resolution, and the other models achieve their best results on medium resolution. This supports our hypothesis that the state-of-the-art BIQA models designed for standard resolution images do not perform optimally on high resolution images. On the other hand, HR-BIQA achieves state-of-the-art performance on full resolution, but since it requires several patches to predict MOS accurately, it performs relatively poorly on low resolution images. 

\section{conclusions}

In this paper, we introduce a new high-resolution image quality database HRIQ, consisting of 1120 images captured in the wild. All the images were rated by at least 24 users in a controlled laboratory environment. We also performed a comprehensive performance evaluation study of different BIQA models on the HRIQ database, using the original database and two downsampled versions of the database with lower resolutions. Our results suggest that even though the state-of-the-art BIQA models can predict low resolution image quality accurately, their performance is not optimal for high resolution input. Substantially better results were obtained by using a patch-based BIQA model designed for high resolution images.

\vfill\pagebreak

\bibliographystyle{IEEEbib}
\bibliography{main.bib}

\end{document}